\documentclass[runningheads]{llncs}
\usepackage[utf8]{inputenc}
\usepackage{mathtools}
\usepackage{amsmath,amssymb,amsfonts}
\usepackage{paralist}
\usepackage{todonotes}
\usepackage{algorithmic}
\usepackage{graphicx}
\usepackage{textcomp}
\usepackage{xcolor}
\usepackage{algorithm2e}
\usepackage{algorithmic}
\usepackage{multirow}
\usepackage[caption=false]{subfig}
\DeclareMathOperator*{\argmin}{arg\,min}
\newcommand{\mymethod}{\textit{CoMT}}

\begin{document}
\title{Robust Federated Training via Collaborative Machine Teaching using Trusted Instances}
%
%
\author{  Yufei Han \inst{1} \and
Xiangliang Zhang\inst{2}}
%
%
\institute{Symantec Research Labs \and King Abdullah University of Science and Technology}
%
\maketitle              
\begin{abstract}
Federated learning performs distributed model training using local data hosted by agents. It shares only model parameter updates for iterative aggregation at the server. Although it is privacy-preserving by design, federated learning is vulnerable to noise corruption of local agents, as demonstrated in previous study on adversarial data poisoning threat against federated learning systems. Even a single noise-corrupted agent can bias the model training. In our work, we propose a collaborative and privacy-preserving machine teaching paradigm with multiple distributed teachers, to improve robustness of the federated training process against local data corruption. We assume that each local agent (teacher)   have the resources to verify a small portions of trusted instances, which  may not by itself be adequate for learning. In the proposed collaborative machine teaching method, these trusted instances guide the distributed agents to jointly select a compact while informative training subset from data hosted by their own. Simultaneously, the agents learn to add changes of limited magnitudes into the selected data instances, in order to improve the testing performances of the federally trained model despite of the training data corruption. Experiments on toy and real data demonstrate that our approach can identify training set bugs effectively and suggest appropriate changes to the labels. Our algorithm is a step toward trustworthy machine learning.
\end{abstract}
\section{Introduction}
The concept of federated learning is proposed by Google in \cite{Konecny2016}. The main idea is to conduct model training using data sets hosted by distributed agents while preventing data leakage. It allows each agent to generate independently local model update with the hosted data instances. The distributed agents only share the local model updates with the central server, where the local updates are averaged to estimate the global drift of model parameters. The federated optimization process is a double-edged sword. For one thing, as no explicit data transfer is conducted, federated learning provides a strong barrier protecting training data privacy. For the other thing, federated optimization can be easily biased by the local noise corruption on any of the agents. Such local noise corruption can appear as outliers that are relatively easy to identify, or as systematic biases. The latter bias usually corrupts the majority of training data, which have much sever adverse effects on learning \cite{Brodley2011}. They are much harder to be detected, because the systematically corrupted data instances appear self-consistent. We propose a novel algorithm to mitigate the adverse impact of systematic noise corruption to achieve robust federated training via {\bf C}ollaborative {\bf M}achine {\bf T}eaching,  thus named {\bf \mymethod}.
At the core of \mymethod, the local \emph{agents} act as distributed \emph{teachers}, while the center \emph{serve} is the \emph{student} co-taught by the distributed \emph{teachers}.
The teachers are organized to jointly select the most informative subset of the hosted distributed data. The data corruption is debugged by a set of ``trusted data instances'' owned by each teacher and verified by domain experts. To minimize the demand of experts' verification, the trusted datasets are usually small in size,  insufficient for learning but important on guiding the selection of trustable instances.  



The collaborative teaching activity and the federated model training are unified within a joint optimization process. It defines an explicit interaction between teaching and learning: the distributed teachers collaborate to generate an appropriate training subset based on the trusted instances. The federated learners train the model on the carefully tuned subset and force the learned model to agree with the trusted data instances throughout the optimization process. The federated teaching-learning process aims to achieve three goals simultaneously. Firstly, the trusted instances guides the model training process despite the systematically corrupted training data. The goal of classic machine teaching methods focus on tuning training data to force the learned model parameter to be as close as possible to the given value. It is far from being practically useful for model training, as we barely know the parameter value of the target model before training. In contrast with the previous machine teaching methods, the proposed \textbf{\textit{CoMT}} algorithm can produce directly an applicable model as the output of robust federated training. Second, the joint optimization of the collaborative teaching and learning process require no explicit data transferring. Therefore the proposed \textbf{\textit{CoMT}} is by design privacy-preserving. Thirdly, the proposed method can provide highly scalable computing over large-scale data sets, whereas previous machine teaching methods have the notorious issue of expensive computational cost. In this sense, the proposed method is more suitable for real-world applications. 

\section{Related Work Discussion and Our Contributions }\label{related_literatures}
\subsection{Machine Teaching}
Machine teaching was originally proposed in~\cite{goldman1995complexity}. Most of the works focus on studying a key quantity called the teaching dimension, i.e., the size of the minimal training set that is guaranteed to teach a target model to the student. For example,~\cite{goldman1995complexity} provides a discussion on the teaching dimension of version space learners,~\cite{liu2016teaching} analyzes the teaching dimension of linear classifiers, and~\cite{zhu2013machine} studies the optimal teaching problem of Bayesian learners. In standard machine teaching, the student is assumed to passively receive the optimal training set generated by teacher. Later works consider other variants of teaching setting, e.g., in~\cite{zilles2011models,balbach2008measuring}, the student and the teacher are allowed to cooperate in order to achieve better teaching performance. More theoretical studies about machine teaching can be found in~\cite{doliwa2014recursive,chen2018understanding,haug2018teaching,liu2017towards,liu2017iterative}. As another popular application, machine teaching can be also used to perform data poisoning attacks of real-world machine learning systems. In such cases, the teacher is viewed as a nefarious attacker who has a specific attack goal in his mind, while the student is some machine learner, then the teaching procedure corresponds to minimally tampering the clean data set such that the machine learner learns some sub-optimal target model on the corrupted training set. Some adversarial attack applications can be found in~\cite{mei2015using,alfeld2016data}.


Instead of artificially designing the training set, \textit{Super Teaching} \cite{ma2018teacher} conducts subset selection over an \textit{i.i.d.} training set. The identified informative subset is then used for model training. 
Mathematically, \textit{Super Teaching} is defined as:
\begin{definition}[Super Teaching]
\label{def:superteaching}
Let $S$ be an $n$-item \textit{iid} training set, and $T$ be a teacher who selects a subset $T(S)\subset S$ as the training subset for learner $A$. Let $\hat\theta_S$ and $\hat\theta_{T(S)}$ be the model learned from $S$ and $T(S)$ respectively. Then $T$ is a super teacher for learner $A$ if~$\forall\delta>0, \exists N$ such that $\forall n\ge N$
\begin{equation}
P_{S}[{R(\hat\theta_{T(S)}) \le c_n R(\hat\theta_S)}] > 1-\delta,
\end{equation}
where $R$ is a teaching risk function, the probability is with respect to the randomness of $S$, and $c_n \le 1$ is a sequence called the super teaching ratio.
\end{definition}

The idea of selecting an informative training subset is also explored in \cite{Yangfan2018}. In the proposed \textit{Learning-to-Teach} framework, the teacher conducting subset selection is modeled with Deep Neural Nets (DNN). The goal of teaching is to select training samples to make faster convergence of the DNN based learner. The teacher network is tuned via reinforcement learning with reward signals encouraging fast descent of the classification loss of the learner. In contrast to learning-to-teach, super teaching in~\cite{ma2018teacher} focuses on a more general teaching goal, which drives the student to learn the expected model. 

Another machine teaching method closely related to our work is \textit{Debugging Using Trusted Items} (DUTI) \cite{XuezhouZhang2018}. \textit{DUTI} assumes that only the labels of the training data are corrupted. Noise over the labels, such as class label flipping noise for classification and continuous-valued noise for regression, is considered as the bug of training data. \textit{DUTI} formulates a two-level optimization problem to conduct the teaching task. It is designed to learn to inject the smallest crafting into the potentially corrupted labels, guided by a small portion of the trusted items. The injected changes are then provided to a domain expert as suggested bug fixes and used to identify candidates of outliers in the training data. The mathematical definition of \textit{DUTI} is given as follows: 

\begin{definition}[DUTI]
\label{def:duti}
Let $(X,Y)$ as the features and labels of training data. $Y$ can be class labels or regression targets. The outliers of the training data corrupt $Y$. $(\tilde{X},\tilde{Y})$ denote the trusted data instances. Let $\mathcal{A}$ be the learning paradigm of the learner, e.g., a regularized empirical risk minimizer with strongly convex and twice differentiable objective function. Conceptually, DUTI solves the two-level optimization problem: 
\begin{equation}
\small
\begin{split}
&\argmin_{Y'} = Distance(Y',Y)\\
&s.t. \,\, Predictor = \mathcal{A}(X,Y'), \,\, Predictor(X) = Y, \,\,  Predictor(\tilde{X}) = Y'\\
\end{split}
\end{equation}
where $Distance$ denotes the distance between the potentially buggy labels $Y$ and the fixed labels $Y'$. $Predictor$ denotes the model learned by fixed training data. In general, \textit{DUTI} finds an alternative labeling $Y'$ for the training data, which stays as close as possible to $Y$. The model trained with the original feature vectors $X$ and the alternative labels $Y'$ correctly predicts the labels $\tilde{Y}$ of the
trusted items $\tilde{X}$, and the alternative labeling is self-consistent.
\end{definition}

Our work is aligned with \textit{Super Teaching} \cite{ma2018teacher} and \textit{Learning to Teach} \cite{Yangfan2018} on selecting training instances without artificially crafting, but involves multiple distributed teachers, rather than a single teacher in \cite{ma2018teacher,Yangfan2018}.
We also face the challenge of systematically corrupted training data $X$ and $Y$, while \textit{DUTI} \cite{XuezhouZhang2018} only addresses the corruption on labels $Y$. 
Again, all these machine teaching study including \textit{DUTI} works in a single teacher setting, while ours focuses on the \textbf{collaborative teaching a set of distributed agents}. We thus discuss the related work of learning in a distributed environment, Federated Learning, in next subsection. Then we will summarize our contribution in subsection \ref{sec:contribution}.

\subsection{Federated Learning}
Federated learning \cite{Konecny2016} is a communication-efficient and privacy-preserving distributed model training method over distributed agents.
Each agent hosts their own data instances and is capable of computing local model update. In each round of model training, the training process is first conducted on each node in parallel without inter-node communication.  
Only the local model updates are aggregated on a centralized parameter server to derive the global model update. The aggregation is agnostic to data distribution of different agents. Neither the centralized server, nor the local agents have visibility of the data owned by any specific agent. In \cite{MJaggi2014,VSmith2018}, a communication-efficient distributed optimization method named \textit{CoCoA} is proposed for training models in a privacy-preserving  way. \textit{CoCoA} applies block-coordinate descent over the dual form of the joint convex learning objective and guarantees sub-linear convergence of the federated optimization. Furthermore, the optimization process does not require to access data instances hosted by each node. Only local dual variable updates need to transfer from local nodes to the central server. This property makes \textit{CoCoA} appropriate for federated training. 

A federated data poisoning attack is recently proposed in \cite{Bhagoji2018}. This work assumes that only one malicious agent conducts non-colluding adversarial data poisoning over the data instances that it hosts. 
Our method is distinct from this work since we study consensus collaboration of multiple teachers in tuning training data. In addition, our method can be used as a data cleaning process to mitigates the effects of malicious noise injection.


\subsection{Our contributions} \label{sec:contribution}
Our work extends the horizon of \textbf{machine teaching} to deliver a \textbf{robust federated learning} scenario. The major contributions of our work can be concluded as in following four aspects. \\
\indent Firstly, unlike the previous machine teaching methods with a single teacher, we organize multiple teachers   as collaborative players in both teaching and learning. Data instances hosted by one teacher cannot be accessed by the others, which prevents the risk of data leakage in the collaboration but also defines a challenging teaching task. Teaching agents can only access their own data, but they are expect to achieve consensus in the joint teaching collaboration. \\ 
\indent Secondly, we assume a more challenging scenario in which over $90\%$ of training data instances are corrupted by systematical noise. The trusted instances are only a small fraction of the potential buggy data. Furthermore, we assume that both features and labels can be noisy. The proposed method can handle a mixtures of both cases with a computationally efficient optimization process. In contrast, \textit{DUTI} only copes with the noisy labels and suffers from the issue of scalability facing large-scale data sets. \\
\indent  Thirdly,  distributed teachers  also learn to change the features of potentially corrupted data with the trusted instances. Enabling more teaching flexibility helps to deliver better teaching performance. \\
\indent Finally, federated teaching and learning in the proposed method are incorporated into a joint distributed optimization problem. The model learned with the tuned training data is produced directly as the output. Coupling teaching and learning together guides the teaching activity with a learning-performance oriented objective. It helps to minimize the changes injected to the training data while still produce satisfying learning performance, in order to reduce the risk of introducing unexpected artefacts into the data. 


\section{Collaborative Machine Teaching using Trusted Instances (CoMT)}\label{problem_defintion}
\subsection{Problem definition}\label{notations}
Assuming that there are $K$ local agents noted as $A_{k}$ (k=1,2,3,...,K), each of them hosts a buggy training set composed by  $\{(X^{k}_i,Y^{k}_i)\}_{i=1:n_{k}}$. We use $X^{k}$ and $Y^{k}$ to denote all the features and labels of one training set. In the setting of this work, both $X^{k}$ and $Y^{k}$ can be potentially contaminated by noise. In addition, we assume that each agent hosts a small portion of trusted data instances $\{(\tilde{X}^{k}_{i},\tilde{Y}^{k}_{i})\}_{i=1:m_{k}}$, where $m_{k} \ll n_{k}$. These trusted instances are verified by domain experts at considerable expense. The labels $Y^{k}$ and $\tilde{Y}^{k}$ can be continuous for regression or discrete for classification. 
The mathematical definition of the proposed collaborative teaching process is given as in Eq.\ref{eq:teachingobj}:
\begin{equation}\label{eq:teachingobj}
\small
\begin{split}
&X',\,\, b^k:k\in[K] = \underset{X', b^{k}:k\in[K]} \argmin \sum_{k=1}^{K}\sum_{i=1}^{n_k}{b^{k}_{i}\|X'^{k}_{i}-X^{k}_{i}\|^2 + \lambda_{b}|b|}\\
s.t. & \,\,\,\, \hat{\theta}^{*} = \underset{\hat{\theta}} \argmin \sum_{k=1}^{K}\sum_{i=1}^{n_{k}} b^k_{i} \ell(\hat{\theta}, X^{'k}_{i},Y^{k}_{i}) + \lambda\Omega(\hat{\theta}), \,\,\,\, b^k = \{0,1\}^{n_k}, \\
& f_{\hat{\theta}^{*}}(\tilde{X}^{k}_i) = \tilde{Y}^{k}_i, \,\,\, f_{\hat{\theta}^{*}}(X'^{k}_{i}) = Y^{k}_{i}\\
\end{split} 
\end{equation}
$\|\|$ measures Euclidean distance between the changed feature $X'^{k}_{i}$ and $X^{k}_{i}$. $b^{k}_{i}$ is a binary indicator denoting whether the corresponding instance is selected for training. $|b|$ denotes L1-norm of $b$.
$\sum_{k,i}\ell(\hat{\theta},X'^{k}_i,Y^{k}_i) + \lambda{\Omega({\hat{\theta}})}$ denotes the joint learning paradigm $\mathcal{A}$ to train the model $f$ parameterized by $\hat{\theta}$. Minimizing Eq.\ref{eq:teachingobj} achieves to simultaneously select an informative subset of training data and inject the minimum changes over the features of the selected data instances. The tuned subset of training data is used to conduct federated training via the learning paradigm $\mathcal{A}$. According to the constraint, the learnt model $f$ should predict consistent labels on both trusted instances and the tuned training data. 

\subsection{Dual form of the collaborative teaching}
The dual objective of the learning paradigm $\mathcal{A}$ for the learner gives: 
\begin{equation}\label{eq:dual_learner}
\small
\alpha^{*} = \underset{\alpha}{\argmin} \sum_{k=1}^{K} \sum_{i=1}^{n_k} \ell^{*}(-\alpha^{k}_i) + \frac{\lambda}{2} \|Z\alpha\|^2 
\end{equation}
where $\ell^{*}$ is the Fenchel dual of the loss function $\ell$. Let $n = \sum_{k=1}^{K} n_k$ denote the total number of training instances owned by the teachers. $Z \in R^{d*n}$ denotes aggregated data matrix with each column corresponding to a data instance. 
The duality comes with the mapping from dual to primal variable: $\omega(\alpha) = Z\alpha$ as given by the KKT optimality condition.    
$\alpha$ is the $N$-dimensional dual variable, where each $\alpha^{k}_i$ denotes the dual variable corresponding to the $i$th data instance hosted by teacher $k$. If $\alpha^{k}_{i}$ diminishes, the corresponding data instance $Z^{k}_{i}$ consequently has no impact over the dual objective in Eq.~\eqref{eq:dual_learner}. Thus, only the data instances with non-zero $\alpha^{k}_i$ dominates the training process. Motivated by this observation, we propose to formulate the objective of the proposed collaborative teaching using the dual form of the learning paradigm in  Eq.\eqref{eq:lr_dual_augmented}.  
\begin{equation}\label{eq:lr_dual_augmented}
\small
\begin{split}
\alpha, Z'  &= \underset{\alpha^{k}_{i},k\in [K],Z'} {\argmin} \frac{1}{n}\sum_{k=1}^{K}\sum_{i=1}^{n_k} \ell^{*}(-\alpha^{k}_i,Z') + \frac{\lambda}{2}\|Z'\alpha\|^2 \\
&+ \lambda_{trusted} \frac{1}{m}\sum_{k=1}^{K}\sum_{i=1}^{m_k}\ell(\tilde{X}^{k}_i, \tilde{Y}^{k}_i,Z'\alpha)  + \lambda_{Z}\|Z'-Z\|^2 + \lambda_{\alpha} \sum_{k=1}^{K}\sum_{i=1}^{m_k} |\alpha^{k}_i|
\end{split}
\end{equation}
where $\lambda_{trusted}$ balances the impact of the trusted data instances in the joint optimization process. Larger $\lambda_{trusted}$ puts more weight over the learning loss of the trusted data instances, which thus sets more strict constraints over the consistency between the learned model and the trusted data. $\lambda_{\alpha}$ is the regularization weight of the adaptive $l_1$-norm penalization enforcing sparsity of $\alpha$ to perform the subset selection. $\lambda_{Z}$ penalizes the magnitudes of the changes injected by the teachers into the buggy training features of the selected subset. An appropriately chosen $\lambda_{Z}$ helps to prevent too much artefacts introduced to the tuned training instances, while still enables the tuning flexibility to deliver efficient teaching. 
The teaching objective given in Eq.~\eqref{eq:lr_dual_augmented} is convex according to the property of Legendre-Fenchel transform. Thus solving Eq.~\eqref{eq:lr_dual_augmented} with gradient descent guarantees fast convergence. As enforced by the $L_1$-norm regularization over $\alpha$, the non-zero entries of $\alpha$ in Eq.~\eqref{eq:lr_dual_augmented} correspond to the selected data instances for the learner to calculate the model parameters. In practice, the learned $\alpha$ has a small fraction of entries with dominant magnitudes, and the rest are negligible. We next demonstrate  how  to  apply  the  proposed  \textit{CoMT}  method  to  two  prevalent  learners, L2-regularized  Logistic Regression (LR) and Ridge Regression (RR). It is worth noting that \textit{CoMT} is not constrained to the two linear models. It is  extendable to federated kernelized learners by introducing random Fourier features \cite{RahRec07}. We leave this extension for future study.

\subsection{CoMT for Ridge Regression and Logistic Regression}
We can instantiate Eq.\ref{eq:lr_dual_augmented} to Ridge Regression by inserting the dual form of Ridge Regression, which gives:
\begin{equation}\label{eq:rr_ruti}
\small
\begin{split}
&\alpha^{*,k},\beta^{*,k} = \underset{\alpha^{k},\beta^{k},k\in [K]}{\argmin}\frac{1}{2\lambda_{w}}\sum_{k=1}^{K}\|(X^{k} + \beta^{k})\alpha^{k}\|^2 + \frac{1}{2}\sum_{k=1}^{K}\|\alpha^{k}\|^2 - \sum_{k=1}^{K}\alpha^{k,T}Y^{k} \\
&+ \lambda_{trusted}\sum_{k=1}^{K}\|\tilde{X}^{k}\frac{1}{\lambda_{w}}(\sum_{k=1}^{K}(X^{k} + \beta^{k})\alpha^{k}) - \tilde{Y}^{k}\|^2 + \lambda_{\alpha}\sum_{k=1}^{K}|\alpha^{k}| + \lambda_{Z}\sum_{k=1}^{K}\|\beta^{k}\|^2\\
\end{split}
\end{equation}
where $\beta^{k}$ denotes the teaching crafting applied to the buggy training data $X^{k}$. $\alpha^{k}\in{R^{n_k}}$ is the dual variable vector. The magnitude of each element in $\alpha^{k}$ measures the contribution of each training data instance hosted by one agent in forming the linear regression parameter $w$. The larger $\alpha^{k}_{i}$ is, the corresponding $X^{k}_i$ and the crafting variable $\beta^{k}_{i}$ will contribute more to recover the regression parameter $w = \frac{1}{\lambda_{w}}\sum_{k=1}^{K}(X^{k}+\beta^{k})^{T}\alpha^{k}$. As shown by Eq.\ref{eq:rr_ruti}, the first three terms enforce the consistency between the learned regression parameter $w$ and the changed training data instances $(X^{k}+\beta^{k})$ and $Y^{k}$. They are derived as the dual definition of Ridge Regression. The forth term is designed to enforce the consistency of the learnt model to the trusted instances $\{\tilde{X}^{k},\tilde{Y}^{k}\}$. 

Similarly, we can define \textit{CoMT} for Logistic Regression in Eq.\eqref{eq:lr_ruti} by combing the dual form of L2-regularized Logistic Regression with Eq.\eqref{eq:lr_dual_augmented}.

\begin{equation}\label{eq:lr_ruti}
\small
\begin{split}
\alpha^{*},\beta^{*} &= \underset{\alpha^{k},\beta^{k},k\in [K]}{\argmin} \sum_{k=1}^{K}\sum_{i=1}^{n_k}(\alpha^{k}_i{\log(\alpha^{k}_i)+(1-\alpha^{k}_i)\log(1-\alpha^{k}_i))} \\
&+ \frac{1}{\lambda_{w}}\|\sum_{k=1}^{K}\sum_{i=1}^{n_k}\alpha^{k}_{i}Y^{k}_{i}(X^{k}_{i} + \beta^{k}_{i})\|^2 \\
&+ \lambda_{trusted}\sum_{k=1}^{K}\sum_{i=1}^{m_k}\log(1+exp(-\tilde{Y}^{k}_{i}(\frac{1}{\lambda_{w}}\sum_{k=1}^{K}\sum_{i=1}^{n_{k}}\alpha^{k}_{i}Y^{k}_{i}(X^{k}_{i}+\beta^{k}_{i}))\tilde{X}^{k}_{i})) \\
&+ \lambda_{\alpha}\sum_{k=1}^{K}|\alpha^{k}| + \lambda_{Z}\sum_{k=1}^{K}\|\beta^{k}\|^2 \,\, \,\,s.t. \,\,\,\, 1\,>\, \alpha^{k}_i\, > \,0 \\
\end{split}
\end{equation}

\section{CoMT optimization}\label{optimization}
We propose to combine Block-Coordinate Descent (BCD) proposed in \cite{VSmith2018} and Alternating Direction Method of Multiplier (ADMM) to solve the optimization problem in Eq.~\eqref{eq:rr_ruti} and Eq.~\eqref{eq:lr_ruti}. In each round of the descent process, we conduct minimization with w.r.t. $\alpha^{k}_{i}$ and $\beta^{k}_i$ belonging to the $k$-th local agent, while fixing all the other $\alpha^{k}$ and $\beta^{k}$. We take the optimization process of \textit{CoMT} for RR for example.Similar steps are applicable to the case of LR. 

We first reformulate Eq.\ref{eq:rr_ruti} into the equivalent form shown in Eq.\ref{eq:rr_ruti_admm}:
\begin{equation}\label{eq:rr_ruti_admm}
\small
\begin{split}
&\alpha^{k,*},\beta^{k,*} = \underset{\alpha^{k},\beta^{k},k\in [K]}{\argmin}\frac{1}{2\lambda_{w}}\sum_{k=1}^{K}\|(X^{k} + \beta^{k})\alpha^{k}\|^2 + \frac{1}{2}\sum_{k=1}^{K}\|\alpha^{k}\|^2 - \sum_{k=1}^{K}{(\alpha^{k})^T}Y^{k} \\
&+ \lambda_{\alpha}\sum_{k=1}^{K}|\alpha^{k}| + \lambda_{Z}\sum_{k=1}^{K}\|\beta^{k}\|^2 + \lambda_{trusted}\sum_{k=1}^{K}\|\tilde{X}^{k}\widetilde{\theta} - \tilde{Y}^{k}\|^2 \\
&s.t. \,\,\,\,\,  \widetilde{\theta} = \frac{1}{\lambda_{w}}(\sum_{k=1}^{K}(X^{k} + \beta^{k})\alpha^{k})
\end{split}
\end{equation}
Following scaled ADMM, we can express Eq.\eqref{eq:rr_ruti_admm} as: 
\begin{equation}\label{eq:rr_ruti_sadmm}
\small
\begin{split}
&\alpha^{k,*},\beta^{k,*} = \underset{\alpha^{k},\beta^{k},k\in [K]}{\argmin}\frac{1}{2\lambda_{w}}\sum_{k=1}^{K}\|(X^{k} + \beta^{k})\alpha^{k}\|^2 + \frac{1}{2}\sum_{k=1}^{K}\|\alpha^{k}\|^2 - \sum_{k=1}^{K}{(\alpha^{k})^T}Y^{k} \\
&+ \lambda_{\alpha}\sum_{k=1}^{K}|\alpha^{k}| + \lambda_{Z}\sum_{k=1}^{K}\|\beta^{k}\|^2 + \frac{\rho}{2}\|\widetilde{\theta} -  \frac{1}{\lambda_{w}}(\sum_{k=1}^{K}(X^{k} + \beta^{k})\alpha^{k}) + u\|^2\\
&\widetilde{\theta}^{*} = \underset{\widetilde{\theta}}{\argmin} \lambda_{trusted}\sum_{k=1}^{K}\|\tilde{X}^{k}\widetilde{\theta} - \tilde{Y}^{k}\|^2 +\frac{\rho}{2}\|\widetilde{\theta} -  \frac{1}{\lambda_{w}}(\sum_{k=1}^{K}(X^{k} + \beta^{k})\alpha^{k}) + u\|^2\\
&u = u + \widetilde{\theta} -  \frac{1}{\lambda_{w}}(\sum_{k=1}^{K}(X^{k} + \beta^{k})\alpha^{k})
\end{split}
\end{equation}
where $\rho > 0$ is the augmented Lagrangian parameter and $u$ is the scaled dual variable of ADMM. The pseudo codes of the optimization procedure is given in Algorithm~\ref{alg:fedML}.

\begin{algorithm}[t]
\caption{Block-Coordinate Descent for CoMT}\label{alg:fedML}
\KwData{Potentially buggy training data $\{X^{k}_{i},Y^{k}_{i}\}_{k = 1,...,K, i = 1,...,n_{k}}$ hosted by $K$ learning agents and the trusted data $\{\tilde{X}^{k}_{i},\tilde{Y}^{k}_{i}\}_{k = 1,...,K, i = 1,...,m_{k}}$ 
} 
\KwIn{$T \geq 1$ as the maximum iteration steps, scaling parameter $1 \leq \gamma_{k} \leq {K}$, by default $\gamma_{k} = 1$. $\rho$ = 1e2 as the augmented Lagrangian multiplier} 
\KwOut{${\alpha^{T,k}_{i}},\beta^{T,k}_{i},k=1,2,...,K,i=1,2,...,n_{k}$}
\textbf{Initialize}: Set $\alpha^{0,k}_{i} = 0$ and $\beta^{0,k}_{i} = 0$ for all $K$ machines. ${\widetilde{\theta}}^{(0)} = 0$ and $u^{0} = 0$\\
\For{$t=1$ \KwTo $T$}{ 
      \For{\textit{all local agents} $k=1,2,3,...,K$ in parallel}{
             $\Delta^{*}{\alpha}^{k},\Delta^{*}{\beta^{k}} = \underset{\Delta{\alpha}^{k},\Delta{\beta^{k}}}{\argmin} 
             \frac{1}{2\lambda_{w}}\|(X^{k} + \beta^{t-1,k} + \Delta{\beta}^{k})(\alpha^{t-1,k} + \Delta{\alpha}^{k})\|^2 + \frac{1}{2}\|\alpha^{t-1,k} + \Delta{\alpha}^{k}\|^2 - (\alpha^{t-1,k} + \Delta{\alpha}^{k})^{T}Y^{k}
             + \lambda_{\alpha}|\alpha^{t-1,k} + \Delta\alpha^{k}| + \lambda_{Z}\|\beta^{t-1,k}+\Delta\beta^{k}\|^2 + \frac{\rho}{2}\|\widetilde\theta^{t-1} - \frac{1}{\lambda_{w}}(X^{k}+\beta^{t-1,k} + \Delta\beta^{k})(\alpha^{t-1,k}+\Delta\alpha^{k}) + u^{t-1}\|^2$\\
             $\alpha^{t,k} = \alpha^{t-1,k} + \frac{\gamma_{k}}{K}\Delta{\alpha}^{k}$\\
             $\beta^{t,k} = \beta^{t-1,k} + \frac{\gamma_{k}}{K}\Delta{\beta}^{k}$
      }
      
      \textit{Reduce on the central parameter server} ${\tau}^{t} = \frac{1}{\lambda_{w}}\sum_{k=1}^{K}\sum_{i=1}^{n_k} \alpha^{t,k}_{i} (X^{k}_{i} + \beta^{t,k}_{i})$
      
      \textit{Broadcast ${\tau}^{t}$ to all $K$ local agents}
      
      \For{\textit{all $K$ local agents} $k=1,2,3,...K$ in parallel}{
          $\Delta{\widetilde{\theta}}^{k} = \underset{\Delta{\widetilde{\theta}}}{\argmin}{\lambda_{trusted}\|\tilde{X}^{k}(\widetilde{\theta} + \Delta{\widetilde{\theta}}^{k}) - \tilde{Y}^{k}\|^2 + \frac{\rho}{2}\|(\widetilde{\theta} + \Delta{\widetilde{\theta}}^{k}) - \tau^{t} + u^{t-1}\|^2}$
      }
      
      \textit{Reduce on the central parameter server} $\widetilde{\theta}^{t} = \widetilde{\theta}^{t-1} + \frac{1}{K}\sum_{k=1}^{K}\Delta{\widetilde\theta}^{k}$ 
      
      \textit{Update on the central parameter server} $u^{t} = u^{t-1} + \widetilde{\theta}^{t} - \tau^{t}$
      
      \textit{Broadcast $\widetilde{\theta}^{t}$ and $u^{t}$ to all $K$ local agents}


    }
\end{algorithm}

We use $\alpha^{t,k}$ and $\beta^{t,k}$ to denote the value of the disjoint block $\{\alpha^k_i\,\beta^{k}_i\}_{i=1,..,n_k}$ estimated at the $t$-th iteration. They correspond to the data instances hosted by the $k$-th local agent. In each round of iteration, we update the dual variables $\alpha^{k}$ and $\beta^{k}$ for each of the $K$ agents in parallel. We assume incremental updates $\Delta\alpha^{k}$ and $\Delta\beta^{k}$ are calculated based on the value of $\alpha^{t-1,k}$ and $\beta^{t-1,k}$. The incremental updates indicate the descent direction minimizing the objective with respect to the block $\alpha^{k}$ and $\beta^{k}$. They are estimated by minimizing the local approximation to Eq.~\eqref{eq:rr_ruti_admm}, where $\alpha^{k}$ is represented as the additional combination $\alpha^{t-1,k} + \Delta{\alpha^k}$. $\gamma_{k}$ is the learning rate adjusting the descent step length for the block $\alpha^{k}$ and $\beta^{k}$. 

In Algorithm.\ref{alg:fedML}, updating each block $\alpha^{k}$ and $\beta^{k}$ does not require the knowledge of the values for the other blocks. All the local updates require  only the values of the variables derived from the last round, $\alpha^{t-1,k}$, $\beta^{t-1,k}$ and the globally aggregated $\widetilde{\theta}^{t-1}$ broadcast from the central server. Similarly, updating $\widetilde{\theta}$ can be conducted using  federated optimization. As such, optimization w.r.t. $\alpha$, $\beta$ and $\widetilde{\theta}$ in Algorithm.\ref{alg:fedML} can be conducted in parallel without inter communication among local agents. 

It is easy to find that:
i)  private data hosted by any local agent is kept on its own device in the collaboration stage. In other words, no training data is transferred directly between agents. Furthermore, updating $\widetilde{\theta}$ only needs to aggregate local updates on the central server to derive  ${\tau}^{t}$ and broadcast it back to the agents. It is difficult to infer any statistical profiles about the training data hosted by local agents solely based on the aggregation $\sum_{i=1}^{n_k} \alpha^{t,k}_{i} (X^{k}_i + \beta^{t,k}_i)$, which prevents the risk of unveiling private data of one local agent to the others in the teaching and learning collaboration. 
ii) Information sharing between local agents is conducted  by updating the global variable $\widetilde{\theta}$ and $\tau$ and then broadcasting the updated value to all $K$ agents in Algorithm.\ref{alg:fedML}. Communication for teaching and learning collaboration is thus efficient, with the cost of $O(Kd+nd)$ in each round of iteration. Moreover, according to \cite{MJaggi2014}, updating $\alpha^{i}$ of local teachers can be triggered with asynchronous parallelism, which allows to organize efficient collaboration of teaching and learning with large number of agents and tight communication budget. Note $\alpha$ is tuned to be a sparse vector. Most entries of $\alpha$ are driven to zeros. We identify the indices of the data instances corresponding to the entries of $\alpha$ with the largest non-zero magnitudes. Only the selected data instances are used to calculate the model parameter. Once $\alpha^{k}$ and $\beta^{k}$ are derived as the output from Algorithm.\ref{alg:fedML}, we can aggregate to obtain the model parameter as $\frac{1}{\lambda_{w}} \sum_{k=1}^{K}\sum_{\tilde{i}\in {S_k}}\alpha^{k}_{\tilde{i}}(X^{k}_{\tilde{i}} + \beta^{k}_{\tilde{i}})$. $S_{k}$ denotes the identified data subset hosted by the local agent $k$. The parameter is calculated by globally ranking of $\alpha$'s entries on the central server and then aggregating local estimation shared by each agent. 

\section{Experimental Results}\label{Experimens}
\subsection{Experimental setup}\label{sec:setup}
We test the proposed \textit{CoMT} algorithm with both synthetic data set and real-world benchmark data sets (summarized in Table.\ref{tab:datasets}). To construct the synthetic dataset, we first create clusters of random data instances following normal distribution, and then assign one half of the clusters as positive and the other half as negative,   to construct a balanced binary-class data set. The regression dataset is obtained by  applying random linear regressor to the created $X$ to get  the regression target $Y$. The dimensionality of each data instance is fixed to $10$. Without loss of generality, we set the number of agents $K$ to 5 in the experimental study. To generate \textit{i.i.d.} data instances, the mean and variance of the normal distribution for data generation are kept the same for different agents. The summary of the real-world data sets is shown by Table.\ref{tab:datasets}, which are used to evaluate practical performances of the proposed method over large-scale real-world data samples. In the empirical study on both synthetic and the real-world data, each local agent is assumed to host the same amount of training instances.  The real-world dataset is randomly shuffled and assigned to each local agent.

In the experimental study over the synthetic data, we first randomly extract 40\% of the whole data set as the training data. These training data are corrupted then to generate buggy training data and assign to all local agents. we choose $\eta{\%}\,(\eta \ll 1)$ of the whole data set as the trusted instances hosted by the local agents. They are considered to be free from noise. The rest of the data are used as testing instances to evaluate the performances of the learned classification or regression model. To construct buggy training sets for local agents, in the case of Ridge Regression, we add random Gaussian distributed noise to both features and regression targets as in Eq.\eqref{eq:noise}: 
\begin{equation}\label{eq:noise}
\small
X_{i} = \hat{X}_{i} + \zeta_{x} * \varepsilon_{x} * \vartheta_{x},\,\,Y_{i} = \hat{Y}_{i} + \zeta_{y} * \varepsilon_{y} * \vartheta_{y}\\
\end{equation}
where $\hat{X}$ and $\hat{Y}$ are the original feature and target of the regression training instances before noise injection. $\zeta_{x}$ and $\zeta_{y}$ are two independently generated random variables, following standardized normal distribution. $\varepsilon_{x}$ and $\varepsilon_{y}$ denotes the averaged magnitudes of the features and targets $\sum_{i}|x_i|$ and $\sum_{i}|y_i|$. $\vartheta_{x}$ and $\vartheta_{y}$ are the magnitudes of the injected noise corruption to each feature vector and target variable. In the case of Logistic Regression, we choose to define two scenarios of buggy training data. First, we fix the class labels of training data, while follow the same protocol in Eq.\eqref{eq:noise} to add Gaussian distributed noise to feature vectors. Second, we fix the training features and randomly flip $40\%$ of the class labels to generate buggy training data for Logistic Regression. To measure the collaborative teaching performances, we repeat the process of random sampling of training and injecting noise for 20 times. The mean and variance of $R$-squared and AUC derived on the testing instances are used as the performance metrics of the regression and classification model. 

\begin{table}[t]
\centering
\small
\caption{Summary of public real-world benchmark datasets.}
\label{tab:datasets}
\resizebox{0.6\linewidth}{!} {
\begin{tabular}{|c|c|c|}
\hline
\textbf{Dataset} & \textbf{No. of Instances} & \textbf{No. of Features} \\ \hline
     IJCNN   &        49,990          &       22           \\ \hline
     CPUSMALL & 8,192 & 12 \\\hline
\end{tabular}
}
\vspace{-0.4cm}
\end{table}

We involve the following baselines to conduct comparative study: 
\begin{itemize}
    \item In both regression and classification case, we use only the trusted instances of each agent to conduct federated model training. The derived models are evaluated on the testing instances. We name it with \textbf{TI-only}. The purpose of involving \textbf{TI-only} is to confirm that learning jointly with both the buggy data and the trusted items helps to achieve better performances. 
    \item We simply the proposed \textit{CoMT} method shown in Eq.\eqref{eq:lr_dual_augmented} by dropping the data change operation. The simplified method, noted as \textbf{CoMT-subset}, only selects subsets for model training and will be compared to evaluate the effectiveness of buggy data correction. Mathematically, it is defined as:
    \begin{equation}\label{eq:lr_dual_subselect}
    \small
    \begin{split}
    \alpha&= \underset{\alpha^{k}_{i},k\in [K]} {\argmin} \frac{1}{n}\sum_{k=1}^{K}\sum_{i=1}^{n_k} \ell^{*}(-\alpha^{k}_i,Z) + \frac{\lambda}{2}\|Z\alpha\|^2 \\
    &+ \lambda_{trusted} \frac{1}{m}\sum_{k=1}^{K}\sum_{i=1}^{m_k}\ell(\tilde{x}^{k}_i, \tilde{y}^{k}_i,Z\alpha) + \lambda_{\alpha} \sum_{k=1}^{K}\sum_{i=1}^{m_k} |\alpha^{k}_i|
     \end{split}
    \end{equation}
    \item \textbf{DUTI} \cite{XuezhouZhang2018} is adapted to the distributed computing scenario, by first running   in a standalone way on each local agent to estimate the  correction of the buggy training data. Then the crafted training data by \textit{DUTI} are sent for the federated model training.  
    \item \textbf{REWEISE} is a robust linear regression method proposed in \cite{Gervini2002}. \textit{REWLSE} is a type of weighted least squares estimator with the weights adaptively calculated from an initial robust estimator. We follow the recommended parameter setting and adopt \textit{REWEISE} on each local agents independently to derive local model updates. The resulted local model updates are aggregated to calculate model parameters globally as in federated learning. 
    \item \textbf{RloR}\cite{JFeng2014} and \textbf{rLR} \cite{JBootkrajang2012} are proposed as a robust logistic regression method against outliers in feature space and label flipping noise, respectively. We use them as baselines in the two scenarios of learning of Logistic Regression with corrupted training data. Since both methods are not designed for distributed computing, we follow the same federated learning protocol as adapting \textit{DUTI} and \textit{REWEISE}.
\end{itemize}
We believe that the keys to the success of the proposed \textit{CoMT} method are two-folds. Firstly, it conducts collaborative machine teaching, which allows to share information of trusted instances and noise patterns among. This helps to jointly tune buggy training data to deliver robust model training. Secondly, it combines seamlessly the data tuning based machine teaching with model training. The combination explicitly defines a learning-performance driven machine teaching, which helps to optimally tune data for better learning. Compared to the proposed method, none of the baseline methods except \textit{CoMT-subset} considers both of the key designs in the collaborative architecture. Therefore we expect the proposed method to deliver a superior model learning performances over the baselines. All the methods  are implemented in Python 2.7 with \textit{Numpy} and \textit{Scipy} packages on a 5-core AWS EC2 public cloud server, with one core per local agent. 

\subsection{Results on synthetic data}
We choose to generate 50000 synthetic data samples and assign them to the local agents, from which we extract and generate buggy training data, trusted instances and testing instances. We vary $\eta{\%}$ from 0.25\%, 0.5\% to 1\% to increase gradually the fraction of trusted instances on each local agent. $\vartheta_{x}$ and $\vartheta_{y}$ are changed from 0.3, 0.6 to 1.2 respectively to simulate increasingly larger magnitudes of noise corruption. Table.\ref{tab:rr} demonstrates the result. The proposed \textit{CoMT} method only needs a subset of the distributed training instance to finally estimate the model parameter. Therefore, in the results produced by \textit{CoMT}, we also record the fraction $\varrho$ of the selected training data instances with respect to the whole training data set. The running time of each method is noted as $\kappa$ and used to evaluate the computational efficiency on the large-scale data set. 

According to the experimental results, the proposed \textit{CoMT} method and its simplified version produce more accurate models compared to the baseline methods involved in the study, given very small portions of trusted instances. Especially when $\eta = 0.1\%$, the proposed \textit{CoMT} and \textit{CoMT-subset} can achieve over 1.7 times R-squared scores and 1.1 times AUC scores compared to the baselines. In both classification and regression scenarios, the baseline methods can also produce better performances with increasing larger $\eta$. The reason is that larger $\eta$ indicates more available trusted instances for training. Nevertheless, the model learnt via \textit{CoMT} and \textit{CoMT-subset} provide consistently superior and stable performances with larger $\eta$. Furthermore, benefited from the subset selection module embedded in the objective function of \textit{CoMT}, the proposed method only needs 70\% on average of the whole training set to conduct training. The capability of subset selection is helpful in the applications when local computing power is limited, as local agents are not obliged to load all the training data instances to compute the model parameters. Compared to \textit{CoMT-subset}, \textit{CoMT} achieves to obtain more accurate model with equal or even less training instances selected. This observation is consistent with our expectation: \textbf{\textit{CoMT} is able to conduct subset selection and crafting simultaneously}. Thus it allows more flexibility to tune training set, which leads to better overall model performances. 

\begin{table*}
\small
\centering
\caption{Comparison of $R$-squared scores in the regression scenario}
\label{tab:rr}
\resizebox{\linewidth}{!} {
\begin{tabular}{|c|c|c|c|c|c|c|c|c|c|c|c|c|c|c|c|c|c|c|}
\hline
\multirow{2}{*}{\textbf{$\vartheta_{x}$}} & \multirow{2}{*}{\textbf{$\eta$}(\%)}  & \multicolumn{4}{|c|}{\textbf{\textit{CoMT}}} & \multicolumn{4}{|c|}{\textbf{\textit{CoMT-subset}}} & \multicolumn{3}{|c|}{\textit{TI-only}} & \multicolumn{3}{|c|}{\textit{DUTI}} & \multicolumn{3}{|c|}{\textit{REWEISE}} \\\cline{3-19}
 &  & $Avg$ & $Var$ & $\varrho$ & $\kappa$ & $Avg$ & $Var$ & $\varrho$ & $\kappa$ & $Avg$ & $Var$ & $\kappa$ & $Avg$ & $Var$ & $\kappa$ & $Avg$ & $Var$ & $\kappa$\\\hline
\multirow{3}{*}{0.3} & 0.10 & \textbf{0.98} & \textbf{1.73e-5} & \textbf{0.70} & \textbf{35.16s} & \textbf{0.95} & \textbf{2.25e-5} & \textbf{0.70} & \textbf{33.50s} & 0.56 & 3.47e-4 & 12.15s & 0.57 & 9.95e-5 & 223s & 0.55 & 4.30e-4 & 20.50s\\\cline{2-19}
 & 0.50 & \textbf{0.98} & \textbf{8.00e-6} & \textbf{0.70} & \textbf{32.20s} & \textbf{0.96} & \textbf{7.43e-6} & \textbf{0.75} & \textbf{33.50s} & 0.89 & 1.00e-4 & 11.50s & 0.85 & 8.37e-4 & 230s &0.82 & 7.93e-4 & 22.20s\\\cline{2-19}
 & 1.00 & \textbf{0.99} & \textbf{1.12e-6} & \textbf{0.65} & \textbf{36.00s} & \textbf{0.96} & \textbf{1.17e-6} & \textbf{0.70} & \textbf{37.00s} & 0.95 & 7.56e-6 & 12.50s & 0.88 & 8.02e-6 & 255s & 0.84 & 1.26e-6 & 22.10s \\\hline 
 \multirow{3}{*}{0.6} & 0.10 & \textbf{0.99} & \textbf{1.03e-6} & \textbf{0.70} & \textbf{39.50s} & \textbf{0.95} & \textbf{1.43e-6} & \textbf{0.75} & \textbf{36.34s} & 0.55 & 4.88e-5 & 15.00s & 0.54 & 5.00e-5 & 289s & 0.57 & 1.25e-4 & 24.30s\\\cline{2-19}
 & 0.50 & \textbf{0.99} & \textbf{1.57e-5} & \textbf{0.70} &\textbf{38.70s} & \textbf{0.95} & \textbf{1.35e-5} & \textbf{0.70} &\textbf{40.20s} & 0.89 & 1.46e-4 & 15.20s & 0.85 & 5.32e-5 & 302s & 0.83 & 8.07e-5 & 23.30s\\\cline{2-19}
 & 1.00 & \textbf{0.99} & \textbf{4.70e-6} & \textbf{0.65} & \textbf{40.55s} & \textbf{0.95} & \textbf{3.50e-6} & \textbf{0.70} & \textbf{39.35s} & 0.96 & 9.82e-6 & 17.00s & 0.89 & 2.04e-5 & 320s & 0.89 & 8.98e-6 & 23.50s \\\hline
 \multirow{3}{*}{1.2} & 0.10 & \textbf{0.99} & \textbf{7.93e-6} & \textbf{0.70} & \textbf{44.50s} & \textbf{0.96} & \textbf{8.20e-6} & \textbf{0.70} & \textbf{44.00s} & 0.54 & 6.30e-5 & 18.45s & 0.56 & 7.55e-5 & 335s & 0.54 & 4.35e-5 & 25.37s\\\cline{2-19}
 & 0.50 & \textbf{0.99} & \textbf{1.92e-6} & \textbf{0.65} & \textbf{45.70s} & \textbf{0.96} & \textbf{1.55e-6} & \textbf{0.70} & \textbf{46.40s} & 0.91 & 6.87e-5 & 18.20s & 0.86 & 2.54e-5 & 368s & 0.85 & 3.05e-5 & 25.04s\\\cline{2-19}
 & 1.00 & \textbf{0.99} & \textbf{5.83e-6} & \textbf{0.70} & \textbf{47.45s} & \textbf{0.96} & \textbf{7.86e-6} & \textbf{0.70} & \textbf{52.50s} & 0.96 & 8.81e-6 & 19.20s & 0.89 & 1.02e-5 & 370s & 0.90 & 2.21e-5 & 26.30s \\\hline
\end{tabular}
}
\vspace{-0.9cm} 
\end{table*}

\begin{table*}
\small
\centering
\caption{Comparison of AUC in the classification scenario I: Feature corruption}
\label{tab:lr1}
\resizebox{0.9\linewidth}{!} {
\begin{tabular}{|c|c|c|c|c|c|c|c|c|c|c|c|c|c|c|c|}
\hline
\multirow{2}{*}{\textbf{$\vartheta_{x}$}} & \multirow{2}{*}{\textbf{$\eta$}(\%)}  & \multicolumn{4}{|c|}{\textbf{\textit{CoMT}}}  & \multicolumn{4}{|c|}{\textbf{\textit{CoMT-subset}}}&\multicolumn{3}{|c|}{\textit{TI-only}}  & \multicolumn{3}{|c|}{\textit{rLR}} \\\cline{3-16}
 &  & $Avg$ & $Var$ & $\varrho$ & $\kappa$ & $Avg$ & $Var$ & $\varrho$ & $\kappa$ & $Avg$ & $Var$ & $\kappa$ & $Avg$ & $Var$ & $\kappa$ \\\hline
\multirow{3}{*}{0.3} & 0.10  & \textbf{0.73} & \textbf{6.06e-4} & \textbf{0.80} & \textbf{67.60s} & \textbf{0.72} & \textbf{4.65e-4} & \textbf{0.80} & \textbf{67.00s} & 0.67 & 7.3e-4 & 32.10s & 0.68 & 4.83e-4 & 48.20s \\\cline{2-16}
 & 0.50 & \textbf{0.88} & \textbf{4.07e-6} & \textbf{0.65} & \textbf{58.50s} & \textbf{0.86} & \textbf{6.16e-6} & \textbf{0.70} & \textbf{59.20s} & 0.83 & 6.19e-6 & 29.84s & 0.73 & 4.65e-6 & 43.00s \\\cline{2-16}
 & 1.00  & \textbf{0.87} & \textbf{6.20e-6} & \textbf{0.65} & \textbf{62.10s}& \textbf{0.86} & \textbf{8.75e-6} & \textbf{0.70} & \textbf{64.00s} & 0.83 & 8.75e-6 & 25.34s & 0.73 & 6.43e-6 & 51.13s \\\hline 
 
 \multirow{3}{*}{0.6} & 0.10  & \textbf{0.86} & \textbf{1.06e-4} & \textbf{0.70} & \textbf{69.20s} & \textbf{0.87} & \textbf{5.53e-5} & \textbf{0.70} & \textbf{72.30s} & 0.82 & 1.30e-3 & 26.21s & 0.79 & 5.53e-4 & 36.10s \\\cline{2-16}
 & 0.50  & \textbf{0.93} & \textbf{2.00e-4} & \textbf{0.70} & \textbf{62.35s} &\textbf{0.93} & \textbf{3.14e-4} & \textbf{0.80} & \textbf{63.05s} & 0.88 & 2.19e-4 & 25.35s & 0.75 & 2.35e-4 & 41.21s \\\cline{2-16}
 & 1.00  & \textbf{0.90} & \textbf{1.57e-4} & \textbf{0.75} & \textbf{59.20s} & \textbf{0.88} & \textbf{1.80e-4} & \textbf{0.75} & \textbf{60.30s} & 0.85 & 5.57e-5 & 25.35s & 0.72 & 3.70e-5 & 51.00s \\\hline 
 
 \multirow{3}{*}{1.2} & 0.10  & \textbf{0.89} & \textbf{4.99e-4} & \textbf{0.70} & \textbf{63.70s} & \textbf{0.88} & \textbf{4.23e-4} & \textbf{0.75} & \textbf{61.25s} & 0.86 & 1.24e-3 & 29.20s & 0.79 & 3.27e-3 & 32.20s \\\cline{2-16}
 & 0.50  & \textbf{0.91} & \textbf{1.53e-5} & \textbf{0.70} & \textbf{62.00s} & \textbf{0.88} & \textbf{1.53e-5} & \textbf{0.70} & \textbf{62.00s} & 0.86 & 2.55e-6 & 30.50s & 0.82 & 2.32e-5 & 38.30s \\\cline{2-16}
 & 1.00  & \textbf{0.85} & \textbf{1.51e-5} & \textbf{0.75} & \textbf{67.10s} & \textbf{0.83} & \textbf{2.54e-5} & \textbf{0.75} & \textbf{72.00s} & 0.81 & 2.12e-5 & 26.44s & 0.77 & 2.61e-6 & 53.00s \\\hline 
\end{tabular}
}
\vspace{-0.9cm} 
\end{table*}

\begin{table*}
\small
\centering
\caption{Comparison of AUC in the classification scenario II: Label flipping noise}
\label{tab:lr2}
\resizebox{\linewidth}{!} {
\begin{tabular}{|c|c|c|c|c|c|c|c|c|c|c|c|c|c|c|c|c|c|c|}
\hline
\multirow{2}{*}{  \textbf{$\vartheta_{y}$}} & \multirow{2}{*}{\textbf{$\eta$}(\%)}  & \multicolumn{4}{|c|}{\textbf{\textit{CoMT}}}  & \multicolumn{4}{|c|}{\textbf{\textit{CoMT-subset}}}& \multicolumn{3}{|c|}{\textit{TI-only}} & \multicolumn{3}{|c|}{\textit{DUTI}} & \multicolumn{3}{|c|}{\textit{RloR}} \\\cline{3-19}
 &  & $Avg$ & $Var$ & $\varrho$ & $\kappa$ & $Avg$ & $Var$ & $\varrho$ & $\kappa$ & $Avg$ & $Var$ & $\kappa$ & $Avg$ & $Var$ & $\kappa$ & $Avg$ & $Var$ & $\kappa$\\\hline
\multirow{3}{*}{0.3} & 0.10  & \textbf{0.77} & \textbf{5.16e-4} & \textbf{0.70} & \textbf{63.20s} & \textbf{0.75} & \textbf{8.68e-4} & \textbf{0.70} & \textbf{62.80s} & 0.72 & 7.30e-4 & 12.00s & 0.63 & 4.58e-5 & 197.32s & 0.69 & 4.52e-5 & 25.30s \\\cline{2-19}
 & 0.50  & \textbf{0.82} & \textbf{3.94e-5} & \textbf{0.65} & \textbf{60.20s} & \textbf{0.83} & \textbf{3.40e-5} & \textbf{0.70} & \textbf{58.35s} & 0.68 & 5.98e-5 & 12.10s & 0.67 & 6.96e-5 & 154.10s & 0.68 & 1.31e-5 & 32.40s \\\cline{2-19}
 & 1.00  & \textbf{0.82} & \textbf{1.30e-4} & \textbf{0.70} & \textbf{58.00s} & \textbf{0.80} & \textbf{1.12e-4} & \textbf{0.70} & \textbf{58.00s} & 0.76 & 2.54e-4 & 15.10s & 0.72 & 2.12e-4 & 198.20s & 0.68 & 4.02e-4 & 27.44s \\\hline 
 
 \multirow{3}{*}{0.6} & 0.10  & \textbf{0.79} & \textbf{5.16e-4} & \textbf{0.70} & \textbf{63.20s}  & \textbf{0.79} & \textbf{3.05e-4} & \textbf{0.70} & \textbf{61.60s} & 0.74 & 7.30e-4 & 12.00s & 0.72 & 4.58e-5 & 197.32s & 0.72 & 4.52e-5 & 25.30s \\\cline{2-19}
 & 0.50  & \textbf{0.86} & \textbf{3.94e-5} & \textbf{0.75} & \textbf{60.20s} & \textbf{0.83} & \textbf{5.09e-5} & \textbf{0.70} & \textbf{62.30s} & 0.80 & 5.98e-5 & 12.10s & 0.77 & 6.96e-5 & 154.10s & 0.76 & 1.31e-5 & 32.40s \\\cline{2-19}
 & 1.00  & \textbf{0.85} & \textbf{1.30e-4} & \textbf{0.70} & \textbf{58.00} & \textbf{0.83} & \textbf{1.93e-4} & \textbf{0.70} & \textbf{60.00} & 0.79 & 2.54e-4 & 15.10s & 0.77 & 2.12e-4 & 198.20s & 0.78 & 4.02e-4 & 27.44s \\\hline 
 
 \multirow{3}{*}{1.2} & 0.10  & \textbf{0.87} & \textbf{5.16e-4} & \textbf{0.70} & \textbf{63.20s} & \textbf{0.87} & \textbf{5.16e-4} & \textbf{0.75} & \textbf{63.00s} & 0.84 & 7.30e-4 & 12.00s & 0.80 & 4.58e-5 & 197.32s & 0.80 & 4.52e-5 & 25.30s \\\cline{2-19}
 & 0.50  & \textbf{0.89} & \textbf{3.94e-5} & \textbf{0.70} & \textbf{60.20s} & \textbf{0.87} & \textbf{2.55e-5} & \textbf{0.70} & \textbf{62.45s} & 0.83 & 5.98e-5 & 12.10s & 0.80 & 6.96e-5 & 154.10s & 0.82 & 1.31e-5 & 32.40s \\\cline{2-19}
 & 1.00  & \textbf{0.87} & \textbf{1.30e-4} & \textbf{0.70} & \textbf{58.00s} & \textbf{0.84} & \textbf{4.57e-4} & \textbf{0.70} & \textbf{57.20s} & 0.83 & 2.54e-4 & 15.10s & 0.82 & 2.12e-4 & 198.20s & 0.82 & 4.02e-4 & 27.44s \\\hline 

\end{tabular}
}
\vspace{-0.7cm} 
\end{table*}

Running \textit{DUTI} with multiple local agents has a significant higher running time. The major bottleneck of \textit{DUTI} is the inverse operation over $n_k$-by-$n_k$ pairwise inner product matrix of the data instances hosted on each local agent. The computational complexity of matrix inverse is O(${n_k}^{3}$), which can be prohibitively expensive to the local devices with limited computing capability. Furthermore, since each local device runs \textit{DUTI} independently with their own data, we can't launch the model training step until the last device finishes running \textit{DUTI} and provides the crafted training data, while the other devices idles. It leads to extra time cost. Compared to \textit{DUTI}, the proposed method costs less than $25\%$ of the running time according to Table.\ref{tab:rr}, Table.\ref{tab:lr1} and Table.\ref{tab:lr2}. In the regression scenario, conducting \textit{CoMT} on 5 local agents requires on average 1500 iterations before converging to stable estimation of $\alpha$ and $\beta$. The classification scenario requires on average 2000 iterations before reaching convergence. The convergence behavior is similar with that reported in \cite{MJaggi2014,VSmith2018}. We can expect faster convergence with smarter settings of learning rates to estimate the optimal incremental update, e.g., nesterov accelerated gradient. 

In addition, we evaluate scalability of the proposed \textit{CoMT} method by increasing the number of all the synthetic data instances from 50,000, 100,000 to 500,000. We fix all the rest settings as described in Section.\ref{sec:setup}. Since scalability is the main focus, we only record the averaged running time derived at the three levels of data volume. 
In the regression scenario, the averaged running time derived at each level of data volume are 36.23s, 79.45s and 384.32s. In the classification scenario, the averaged running time derived corresponding to each data volume are 65.60s, 141.88s and 684.33s respectively. The results demonstrate approximately linear increasing of computational cost. It verifies the computational efficiency of the consensus optimization in Algorithm.\ref{alg:fedML}. 

\subsection{Results on real-world data sets}
Two real-world data sets, \textit{CPUSMALL} and \textit{IJCNN}, are employed to evaluate practical usability of \textit{CoMT} for $L_2$-regularized Logistic Regression and Ridge Regression respectively. We fix $\vartheta$ globally as 1.5. $\eta$ is set as $5\%$ for regression and $1\%$ for classification respectively. We repeat the sampling process for 20 times and calculate the mean and variance of the derived R-squared scores and AUC scores to measure the performance of \textit{CoMT}. On the regression data set \textit{CPUSMALL}, \textit{CoMT} selects only $55\%$ of the training instances hosted by all local agents to achieve its highest AUC score of 0.71. In contrast, the averaged R-squared score of \textit{TI-only}, \textit{DUTI} and \textit{rLR} are 0.38, 0.56 and 0.58.  On the classification set \textit{IJCNN}, AUC score of \textit{CoMT} achieves 0.72 in the scenario where only the features are corrupted. With the sames setting, AUC scores of \textit{TI-only}, \textit{DUTI} and \textit{RloR} are 0.70, 0.68 and 0.69. In the scenario where only label flipping noise is witnessed, \textit{CoMT} obtains an AUC score of 0.73. Compared to \textit{CoMT}, AUC scores of all the other baseline methods are less than 0.70. In the classification scenario, \textit{CoMT} achieves the best AUC with $55\%$ of the whole training data. After reaching the highest R-squared and AUC scores, it is interesting to observe the performances of the learnt regression and classification model do not increase consistently or even decline with more training data instances selected and crafted. This observation confirms empirically the existence of the optimal subset for the collaborative teaching activity.

\begin{figure}[t]
    \subfloat[CPUSMALL] {
        \raisebox{-\height}{\includegraphics[width=0.45\linewidth]{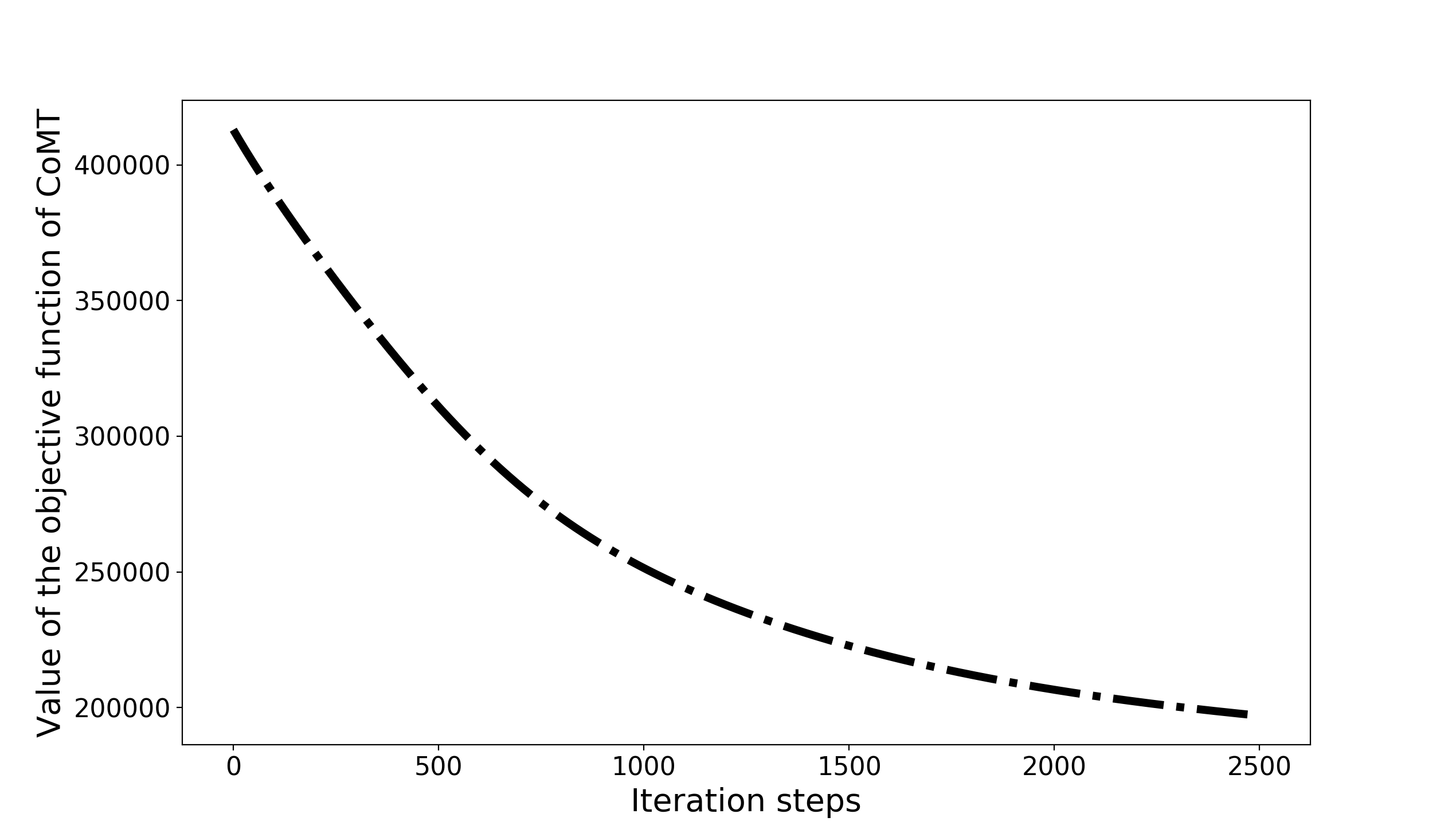}\label{fig:cpu_obj}}
    }
    \hfill
    \subfloat[IJCNN]{
        \raisebox{-\height}{\includegraphics[width=0.46\linewidth]{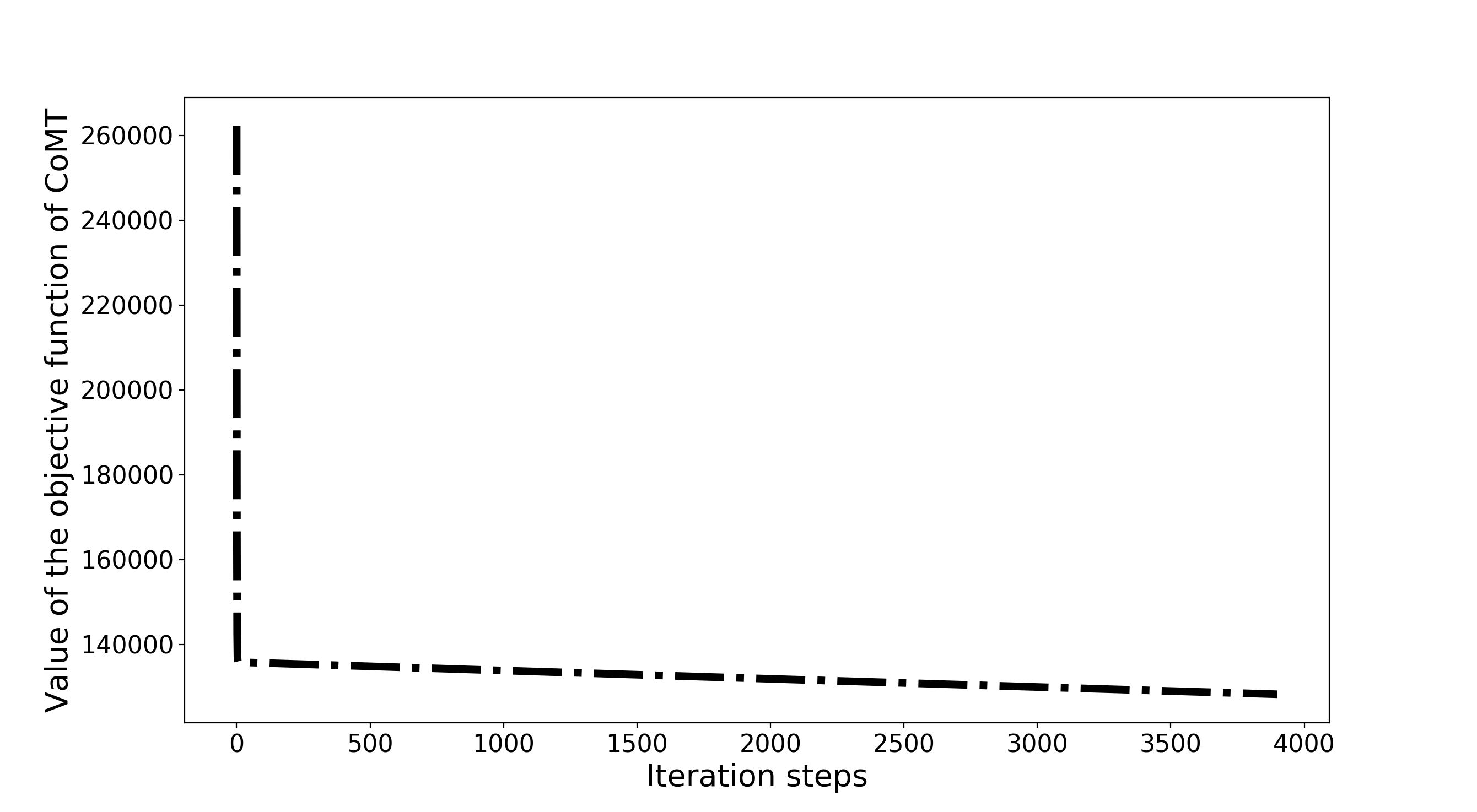}\label{fig:ijcnn_obj}}
    }
    \hfill
    \caption{Convergence of \textit{CoMT} on \textit{CPUSMALL} and \textit{IJCNN} data sets}
	\label{fig:objectivefunc}
\end{figure}

Figure~\ref{fig:cpu_obj} shows that the objective function value of the proposed \textit{CoMT} method converges after 2500 iterations of the collaborative optimization. In this experiment, it costs 45.48s on the given computing platform. Similarly, Figure~\ref{fig:ijcnn_obj} illustrates the declination of the objective functions values on \textit{IJCNN}. In the classification scenarios, \textit{CoMT} requires 1200 iterations before reaching the stable estimation of the model parameter, which cost 195.48s.

%
%
\section{Conclusion}\label{conclusion}
In this paper, we explore how to produce robust federated model training with systematically corrupted data sets distributed over multiple local agents. To attack this problem, we propose a consensus and privacy-preserving optimization process which unifies collaborative machine teaching and model training together. Our main contributions can be concluded in two major aspects. Firstly, tuning of training data and learning with the tuned data are unified together as a joint optimization problem. It helps to better tune buggy training data to make the learnt model consistent with the underlying true correlation between features and labels. Secondly, collaboration between local agents shares information about data tuning, which is used to jointly generate the crafted training data to achieve the teaching goal. Our empirical results on both synthetic and real-word data sets confirm the superior performances of the proposed method over the non-collaborative solutions. Our future work studies practical use of the proposed robust federated training framework over more complex machine learning models. More concretely, we plan to extend the teaching paradigm to diverse types of machine learning models, like deep neural nets.

\bibliographystyle{splncs04}
\bibliography{ijcnn}
\end{document}